\documentclass[10pt,twocolumn,letterpaper]{article}
\PassOptionsToPackage{table}{xcolor}
\usepackage[pagenumbers]{cvpr}
\def\paperID{*****} % Replace with the actual CVPR 2027 paper ID when assigned.
\def\confName{CVPR}
\def\confYear{2027}

\usepackage{amsmath,amsfonts,bm}

\def\eqref#1{equation~\ref{#1}}
\def\1{\bm{1}}

\def\ra{{\textnormal{a}}}

\def\rx{{\textnormal{x}}}

\def\rva{{\mathbf{a}}}

\def\erva{{\textnormal{a}}}

\def\ervx{{\textnormal{x}}}

\def\rmA{{\mathbf{A}}}

\def\vmu{{\bm{\mu}}}
\def\vtheta{{\bm{\theta}}}
\def\va{{\bm{a}}}

\def\ve{{\bm{e}}}

\def\vx{{\bm{x}}}

\def\eva{{a}}

\def\mA{{\bm{A}}}

\def\mH{{\bm{H}}}
\def\mI{{\bm{I}}}
\def\mJ{{\bm{J}}}

\def\mX{{\bm{X}}}

\def\mSigma{{\bm{\Sigma}}}

\DeclareMathAlphabet{\mathsfit}{\encodingdefault}{\sfdefault}{m}{sl}
\SetMathAlphabet{\mathsfit}{bold}{\encodingdefault}{\sfdefault}{bx}{n}
\newcommand{\tens}[1]{\bm{\mathsfit{#1}}}
\def\tA{{\tens{A}}}

\def\tX{{\tens{X}}}

\def\gG{{\mathcal{G}}}

\def\sA{{\mathbb{A}}}
\def\sB{{\mathbb{B}}}

\def\sS{{\mathbb{S}}}

\def\emA{{A}}

\newcommand{\etens}[1]{\mathsfit{#1}}

\def\etA{{\etens{A}}}

\newcommand{\E}{\mathbb{E}}

\newcommand{\R}{\mathbb{R}}

\newcommand{\KL}{D_{\mathrm{KL}}}
\newcommand{\Var}{\mathrm{Var}}

\newcommand{\Cov}{\mathrm{Cov}}
\newcommand{\normltwo}{L^2}
\newcommand{\normlp}{L^p}

\newcommand{\parents}{Pa} % See usage in notation.tex. Chosen to match Daphne's book.

\usepackage{amssymb}
\usepackage{graphicx}
\usepackage{booktabs}
\usepackage[table]{xcolor}
\usepackage{array}
\usepackage{microtype}

\definecolor{cvprblue}{rgb}{0.21,0.49,0.74}
\usepackage[pagebackref,breaklinks,colorlinks,allcolors=cvprblue]{hyperref}
\usepackage{url}

\newcommand{\best}[1]{\textbf{#1}}
\newcommand{\myparagraph}[1]{\noindent\textbf{#1}}
\newcommand{\methodname}{AVE}
\newcommand{\datasetname}{AVE-Dataset}
\newcommand{\opSwap}{\emph{Swap}}
\newcommand{\opInsert}{\emph{Insert}}
\newcommand{\opDelete}{\emph{Delete}}
\newcommand{\syncrope}{Modality-Aware Sync-RoPE}

\title{Edit in Sync: Unified End-to-End Audio-Visual Editing for Human-Centric Videos}

\author{Ziye Yuan$^{1,2,5,*,\ddagger}$\quad Minghong Cai$^{2,*}$\quad Xu He$^{3,*}$\quad Ruchang Yao$^{2}$\quad Xiao Fu$^{2}$\\
Kaixin Ding$^{4}$\quad Ming Zhang$^{1}$\quad Xiangyu Du$^{5}$\quad Tianpeng Lan$^{5}$\quad Da Xie$^{5}$\quad Xiangyu Yue$^{2,\dagger}$\\[3pt]
{\small $^{1}$Peking University\quad $^{2}$MMLab, The Chinese University of Hong Kong\quad $^{3}$Tsinghua University}\\
{\small $^{4}$The University of Hong Kong\quad $^{5}$Kling AI}}

\makeatletter
\patchcmd{\@maketitle}{\vskip .375in}{\vskip .125in}{}{\PackageError{paper}{Title spacing patch failed}{}}
\patchcmd{\@maketitle}{\vspace*{24pt}}{\vspace*{12pt}}{}{\PackageError{paper}{Author spacing patch failed}{}}
\g@addto@macro\@maketitle{%
\vspace{-18pt}% Set the gap between the author block and the full-width teaser.
\begin{minipage}{\textwidth}
   \centering
   \includegraphics[width=\linewidth]{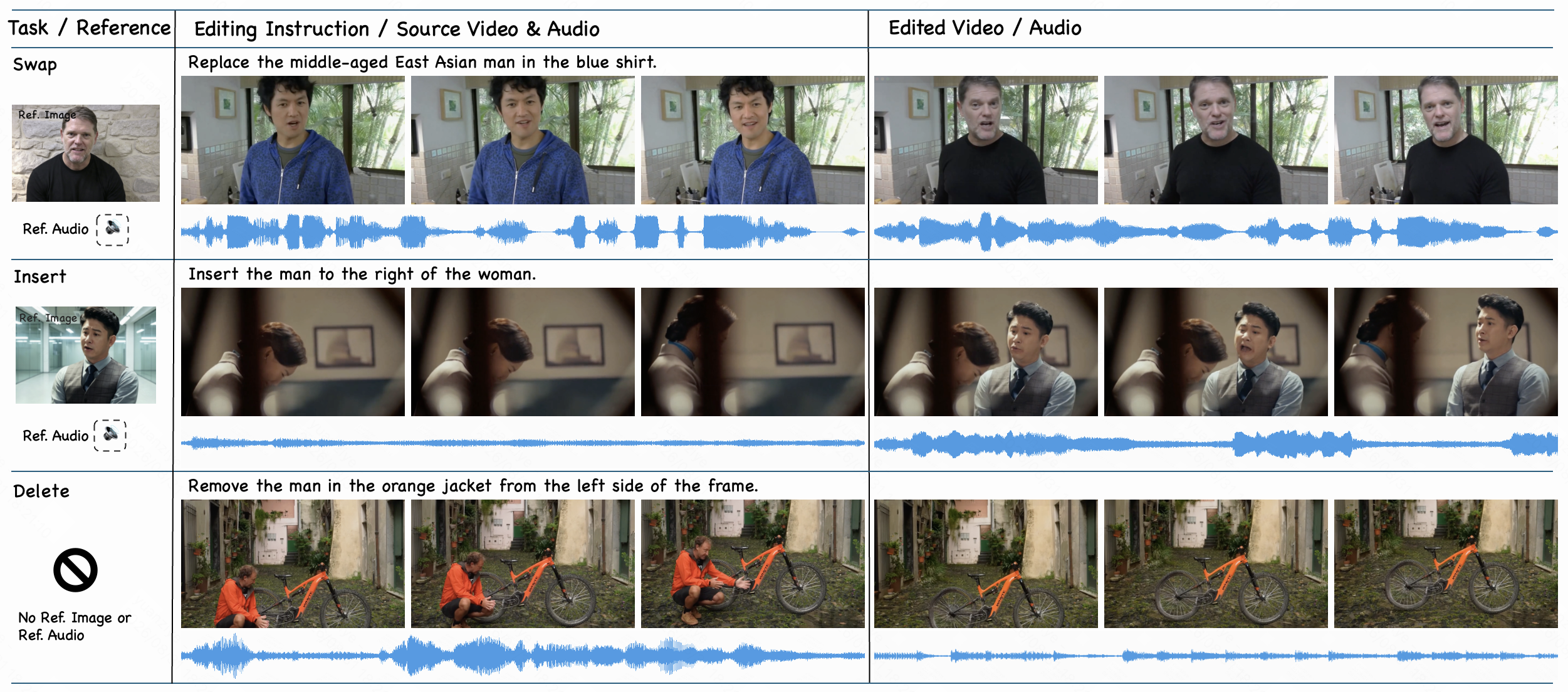}
   \captionsetup{hypcap=false}\captionof{figure}{Overview of \methodname{} across person swapping, insertion, and deletion. Appearance and voice references are provided for swapping and insertion, while deletion requires no reference.}
   \label{fig:teaser}
\end{minipage}\par\vspace{\baselineskip}
}
\makeatother

\begin{document}

\maketitle

\begingroup
\renewcommand{\thefootnote}{\fnsymbol{footnote}}
\footnotetext[1]{Equal contribution.\quad $^{\ddagger}$MMLab intern; work done at Kling AI.}
\footnotetext[2]{Corresponding author.}
\endgroup

\begin{abstract}
Human-centric audio-visual editing requires precise alignment of lip movements, gestures, and expressions with speech.
Existing methods process inputs through independent video and audio pipelines, preventing cross-modal interaction between visual editing and audio dubbing. Many of these pipelines further rely on mask-based inpainting. As a result, they suffer from poor lip-audio synchronization, cumulative pipeline errors, and an inherent mask-size dilemma.
% We present \methodname{}, the first end-to-end framework that unifies all three editing tasks (\opSwap{}, \opInsert{}, and \opDelete{}) guided by reference images and audio within a single  architecture.
We present \methodname{}, the first unified end-to-end framework that jointly performs reference-guided person swapping, insertion, and deletion across both audio and video modalities without requiring spatial masks.
%To overcome the scarcity of paired editing data, we design a scalable synthesis pipeline that constructs high-quality audio-visual editing pairs.
To overcome data scarcity, we design a synthesis pipeline and introduce \datasetname{}, the first audio-visual editing dataset to provide 100K pairs with appearance and voice references.
% 强调一下现在确实没有这样的数据对（限定在pair且有refeerence），也没这么大规模
By combining Synthetic-to-Real reconstruction with preliminary audio warm-up, our framework transcends synthetic pipeline limits and mitigates audio-video token imbalance.
\methodname{} outperforms modality-separate audio-visual editing pipelines in editing realism, temporal consistency, and audio-visual alignment.
\end{abstract}

\section{Introduction}

% 第一段：现有单独的视觉内容编辑（举例xxx）；音频内容编辑的方法； -> 现实世界更需要多模态的协同 -> 引入音视频联合编辑任务 
Content editing has long been an important topic in both visual and audio fields.
Representative visual editing tasks include object removal~\citep{zhou2023propainter,zi2026minimax,fu2026effecterase,miao2026rose}, insertion~\citep{bai2024anything,tu2025videoanydoor,zhao2025dreaminsert,jin2025insertanywhere}, and replacement~\citep{Pan2025RASA,Fu2025ObjectAVEdit,Xu2026MoCha}, while audio editing commonly involves voice conversion~\citep{li2023freevc,Liu2024SeedVC}, source separation~\citep{subakan2021attention,liu2024separate,Lu2024BSRoFormer}, and speech content editing~\citep{peng2024voicecraft,ren2026edit}.
%
% Despite progress in both modalities, independent audio-visual editing remains insufficient for realistic content creation, as visual events are intrinsically coupled with their acoustic counterparts~\citep{chen2024soundingactions,liu2024syncflow}.
% %
% Such interdependence is particularly pronounced in human-centric scenes, where speech is tightly coupled with visual cues including lip movements, facial expressions, hand gestures, and body motion~\citep{lin2024cyberhost,He2025XDub}.
%
Despite progress in both modalities, independent audio-visual editing remains insufficient for human-centric scenarios where visual events and speech are intrinsically coupled across lip movements, facial expressions, and bodily actions~\citep{lin2024cyberhost,He2025XDub}.
Human-centric editing such as insertion, removal, and replacement should be reflected coherently in both modalities, with consistent subject presence, identity, and timing~\citep{Guo2026DreamIDOmni,Liang2026SpongeBob}.
Thus, joint human-centric audio-visual editing has emerged as a key multimodal frontier and the primary focus of this work.

% 第二段：现有的联合编辑， 级联方式，存在误差累积等问题； （收到现有音视频联合生成基模发展的启发）我们希望做一个joint编辑的统一框架。
\begin{figure}[t]
   \centering
   \includegraphics[width=\linewidth]{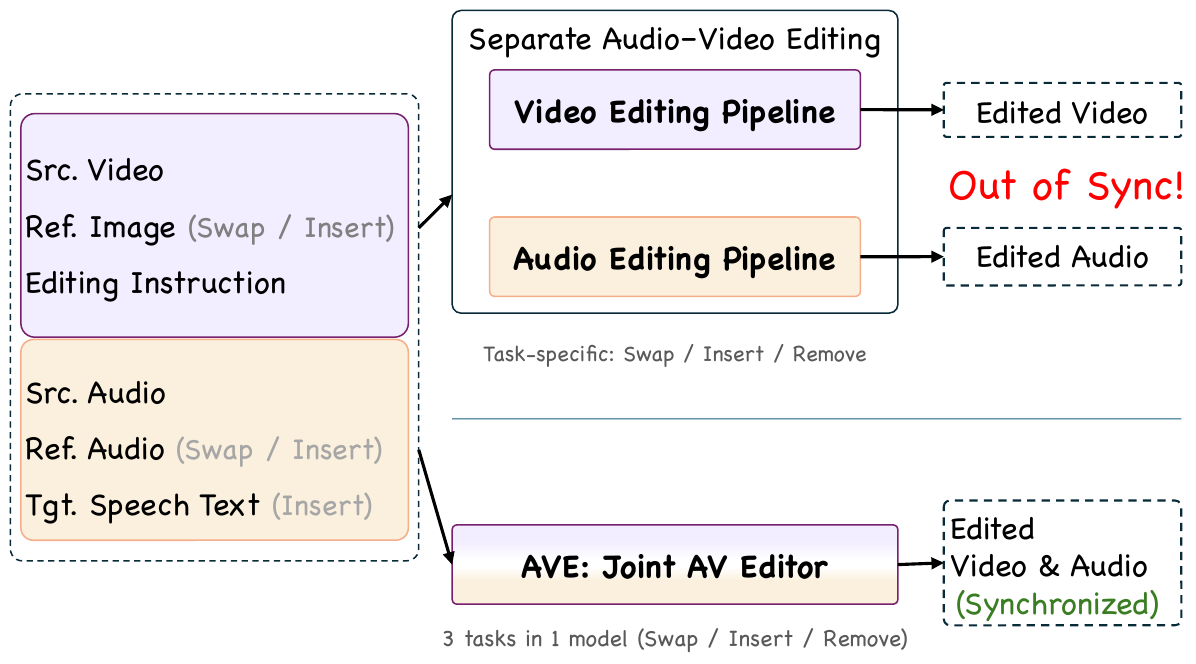}
   \caption{Separate editing pipelines versus \methodname{}}
   \label{fig:method-comparison}
\end{figure}
Existing approaches to audio-visual editing typically decouple video and audio processing into independent pipelines~\citep{Ishii2025CoherentAVEdit,Wu2025VAInpaint}.
Consequently, the visual editor remains speech-agnostic while the audio stage lacks visual guidance; this absent reciprocal conditioning degrades synchronization and compounds pipeline errors (Fig.~\ref{fig:method-comparison}).
Emerging joint audio-video generators prove that unified cross-modal modeling naturally leads to more synchronized and realistic outputs~\citep{Low2025Ovi,HaCohen2026LTX2}.
Inspired by these advances, we develop an end-to-end framework that unifies the core suite of human-centric operations into a single model.

% 第三段：任务的难点在数据； 现有的方法，逃避数据问题，回退到一个抹除后自重建的方式； 我们的思路是构建一个数据管线，把数据的贡献展开说；（我觉得不要强调mask 包括我们的方法也不要强调mask-free 其实说end2end是不是一定程度上就可以认为是mask-free的了）
% However, training such a unified framework is severely constrained by data scarcity: end-to-end supervision requires paired pre- and post-edit audio-visual data, which is virtually non-existent in the wild and prohibitively costly to collect manually at scale.
However, training such a unified framework is severely constrained by data scarcity. End-to-end supervision requires paired audio-visual editing data, which is virtually non-existent in the wild and costly to collect manually at scale.
A common workaround is to reformulate editing as a self-supervised mask-and-reconstruct task~\citep{Cheng2025WanAnimate,Guo2026DreamIDOmni,Liang2026SpongeBob}, where the target subject and speech are masked to reconstruct the original content from the remaining context.
%
% Although scalable to unlabeled in-the-wild data, mask-dependent training forces strict spatial constraints that often erase fine motion cues and surrounding context, causing visual artifacts and temporal instability.
Yet, such mask-dependent training imposes strict spatial constraints that often erase fine motion cues and surrounding context, causing visual artifacts and temporal instability.
%
% To overcome this bottleneck, we build a scalable data construction pipeline that leverages specialized visual and audio models to synthesize high-quality editing pairs from in-the-wild videos.
To overcome this bottleneck, we build a three-stage data construction pipeline.
We leverage specialized visual and audio models to synthesize diverse editing pairs.
%
% To provide reliable visual and audio references, we primarily retrieve cross-paired samples of the target identity from other scenes. We further complement this with targeted data augmentation, ensuring references preserve identity and timbre while varying pose, background, and speech content to prevent copy-and-paste shortcuts.
% Furthermore, we construct cross-pair multimodal references by retrieving the same identity across distinct scenes and applying targeted augmentations, providing reliable identity and timbre guidance while preventing copy-and-paste shortcuts.
To improve reference adherence across diverse visual and acoustic contexts, we construct cross-pair multimodal references by retrieving the target identity across distinct scenes or synthesizing it with generative augmentations, providing reliable identity and timbre guidance while preventing copy-and-paste shortcuts.
Through multi-stage visual and acoustic filtering, we assemble \datasetname{}, comprising 100K curated audio-visual pairs that support unified character \opDelete{}, \opInsert{}, and \opSwap{}.

% 第四段，再说利用这些数据，我们就可以训练一个端到端的统一框架实现这样一个任务；展开一点；   为了进一步解决数据可能存在的问题（为了超越数据质量），我们使用 multi-stage training curriculum
% Based on \datasetname, we propose \methodname, which dispenses with explicit masks entirely. 
Based on \datasetname, we propose \methodname{}, an end-to-end framework for human-centric joint audio-visual editing, supporting three tasks within a unified architecture.
By taking complete source audio-video signals directly as input \methodname{} preserves full spatiotemporal and acoustic context, producing harmoniously synchronized output under multimodal conditions.
\methodname{} builds on a joint audio-video backbone~\citep{Low2025Ovi}, utilizing its bidirectional cross-modal attention for synchronized editing across modalities.
To handle complex multimodal inputs, we introduce a \syncrope{} design that assigns role-specific positional coordinates to source, target, and reference tokens.
This preserves source-target temporal correspondence while aligning audio and video along a shared timeline.
Furthermore, we introduce a multi-stage training curriculum with bidirectional supervision: forward supervision (from a real source to a synthetic target) teaches the intended editing operations, while the backward direction exposes the model to real target distributions.
Together with audio warm-up, bidirectional joint adaptation, and quality-focused refinement, this curriculum mitigates synthesis artifacts and enables the mask-free editor to bootstrap beyond the quality ceiling of its synthetic training data.

% P5 总结 
Our main contributions are as follows:
(1) We present \methodname{}, a unified mask-free framework for human-centric audio-visual editing. It employs \syncrope{} to enable comprehensive cross-modal interaction while preserving audio-visual temporal correspondence.
% (1) We present \methodname{}, an end-to-end, mask-free framework for joint audio-visual editing in diverse human-centric talking-video scenarios, supporting character insertion, deletion, and replacement within a unified architecture. It employs bidirectional cross-modal attention and a \syncrope{} design to enable comprehensive cross-modal interaction while preserving audio-visual temporal correspondence.
%新任务
(2) We develop a scalable data synthesis pipeline to construct \datasetname{}, providing 100K verified audio-visual editing pairs equipped with appearance and voice references alongside a standardized benchmark.
% 数据
(3) We introduce a multi-stage training curriculum with bidirectional supervision, enabling the final model to progressively surpass the quality ceiling imposed by synthetic training data.
% +高效训练，训练方案。

\section{Related Work}

% 1. Joint Audio-Video Generation and Editing
Joint audio--video foundation models learn visual and acoustic content as a coupled distribution rather than separate streams. Early works explored coupled U-Nets for unconditional synthesis~\citep{Ruan2023MMDiffusion}, while recent DiT-based models scale cross-modal generation to open domains with strong multimodal priors~\citep{Low2025Ovi,HaCohen2026LTX2,SkyReelsTeam2026SkyReelsV4}. However, general generation capability alone does not naturally resolve the central editing problem: transferring requested attributes while preserving operation-irrelevant source content. Initial editing formulations adapted single-modality models to paired image--audio data~\citep{Liang2024LanguageGuided} or relied on test-time optimization on video~\citep{Lin2026AvED}. Modular pipelines follow a decomposition that separately inverts, edits, or dubs audio and video~\citep{Fu2025ObjectAVEdit,Wu2025VAInpaint,Ishii2025CoherentAVEdit,Zheng2025AVIEdit}, yet these one-way cascades lack bidirectional interaction and inevitably compound pipeline errors. Recent works move toward coupled end-to-end architectures~\citep{Xu2025SAVE,Zheng2026InstructAV2AV,Chen2026JAVEdit,Liang2026SpongeBob}, yet they remain restricted to single operations~\citep{Xu2025SAVE} or rely on mask-guided routing~\citep{Liang2026SpongeBob}. In contrast, we combine external appearance and timbre references with mask-free localization to unify person-level \opSwap{}, \opInsert{}, and \opDelete{} within a single generative process.

% 2. Reference-Conditioned Human-Centric Generation and Editing
In human-centric scenarios, existing studies have separately investigated visual identity transfer~\citep{Pan2025RASA,Xu2026MoCha} and speech-driven facial animation~\citep{Manela2025EdiDub,He2025XDub}. While achieving promising portrait fidelity and lip synchronization, they primarily operate on the visual stream or require external driving audio, failing to jointly transfer appearance and voice. Conversely, joint dubbing and audio-video personalization methods couple appearance and voice for identity-consistent generation~\citep{Chen2026JustDubIt,Li2026OmniCustom,Chen2026IdentityPresence,Dahan2026IDLoRA,Chen2026OmniCustomizer,Li2026MMControl}, but they synthesize new scenes from scratch rather than preserving source-video context. DreamID-Omni~\citep{Guo2026DreamIDOmni} represents the closest effort by supporting reference-based generation and person replacement. Nonetheless, it does not unify person insertion and deletion under a consistent editing framework, nor does it preserve source motion and linguistic context as editing invariants. In contrast, our approach enables holistic identity and timbre control across all three operations.

% 3. Audio Editing
Audio editing provides essential capabilities for manipulating acoustic content in human-centric scenes. Zero-shot voice cloning synthesizes speech in an arbitrary timbre for character insertion~\citep{Du2024CosyVoice,Qin2023OpenVoice}, while voice conversion preserves linguistic timing and prosody for character replacement~\citep{Liu2024SeedVC}. In parallel, speech separation and extraction isolate target voices from acoustic mixtures to facilitate character deletion~\citep{shi2025sam,Wu2025VAInpaint}. However, these audio-centric methods cannot exploit visual cues to resolve speaker attribution and lip synchronization in talking videos, nor can separation alone reconstruct vacated visual regions. Our framework integrates these editing capabilities into a unified multimodal backbone, allowing acoustic and visual decisions to co-evolve seamlessly.

\section{Method}
\label{sec:problem_formulation}
We study instruction-guided audio-visual editing for talking videos across three person-level operations: \opSwap{}, \opInsert{}, and \opDelete{}. Given a source video $\mathbf{v}_s$, its synchronized audio $\mathbf{a}_s$, a natural-language instruction $c$, and optional multimodal references $(\mathbf{r}_v, \mathbf{r}_a)$ providing the target appearance and vocal timbre (with $\mathbf{r}_v, \mathbf{r}_a = \varnothing$ for \opDelete{}), our goal is to predict the edited audio-visual pair $(\hat{\mathbf{v}}, \hat{\mathbf{a}})$ using a single unified model $f_\theta$:
\begin{equation}
    (\hat{\mathbf{v}},\hat{\mathbf{a}})
    = f_{\theta}(\mathbf{v}_s,\mathbf{a}_s,\mathbf{r}_v,\mathbf{r}_a,c),
    \label{eq:task}
\end{equation}
% where $(\mathbf{r}_v,\mathbf{r}_a)$ is supplied for \opSwap{} and \opInsert{} and set to $\varnothing$ for \opDelete.
% where $(\hat{\mathbf{v}},\hat{\mathbf{a}})$ denotes the predicted synchronized audio-video pair, and $(\mathbf{r}_v,\mathbf{r}_a)$ is set to $\varnothing$ for \opDelete{}. 
The core challenge lies in editing both modalities jointly without explicit spatial signals: $f_{\theta}$ must implicitly resolve the target referent from $c$, maintain precise lip-speech alignment across modalities, and faithfully preserve non-target context.

% Building on this formulation, we propose \methodname{}, a unified, end-to-end, mask-free framework in which a single model performs joint audio-visual editing.
% To address these challenges, we present \methodname{}. In this section, we first describe our scalable data synthesis pipeline for obtaining complex editing pairs in \textbf{Sec.~\ref{sec:data}}. We then present our editing framework, adapted from a joint audio-video generation backbone through a \syncrope{} design tailored to the roles of source, target, and reference tokens, in \textbf{Sec.~\ref{sec:framework}}. Finally, we introduce a progressive, multi-stage training curriculum with bidirectional supervision that lets \methodname{} surpass the quality ceiling of its synthetic training data in \textbf{Sec.~\ref{sec:training}}.
Building on this formulation, we present \methodname{}. In this section, we first describe our data synthesis pipeline for obtaining complex editing pairs in \textbf{Sec.~\ref{sec:data}}. Then we present our editing framework with \syncrope{} design in \textbf{Sec.~\ref{sec:framework}}. Finally, Sec.~\ref{sec:training} introduces a progressive training curriculum with bidirectional supervision to overcome the quality ceiling of synthetic training data.

\subsection{Data}
\label{sec:data}

\begin{figure*}[t]
    \centering
    \includegraphics[width=0.97\linewidth]{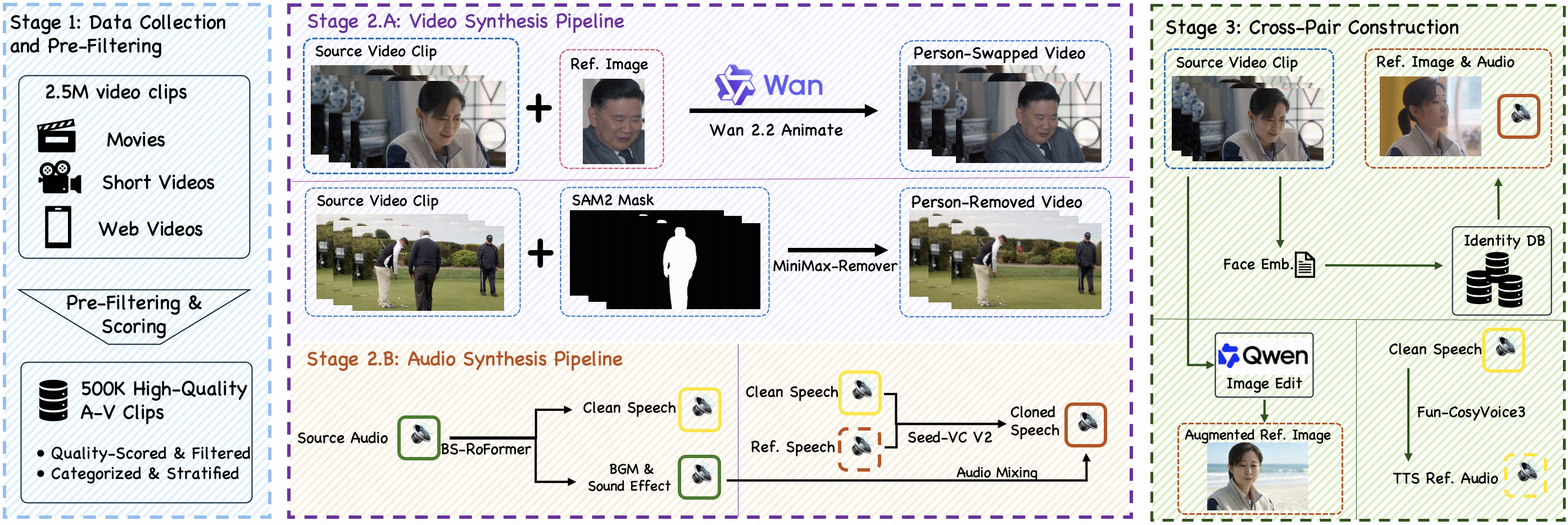}
    \caption{Automated data construction pipeline of \datasetname. The workflow consists of three phases: (1) raw corpus collection and quality pre-filtering, (2) video-audio synthesis for editing pair creation, (3) identity retrieval and multimodal augmentation for cross-pair reference construction.}
    \label{fig:data}
\end{figure*}

% 总述与规模，然后写清楚每一段讲了什么
To obtain scalable supervision for joint audio-visual editing, we build an automatic data construction pipeline illustrated in Fig.~\ref{fig:data}. In Phase 1, we filter 2.5M raw video clips to retain a 500K high-quality, dialogue-rich source corpus (see Appendix~\ref{app:source-prefilter} for details). In Phase 2, we synthesize paired examples for \opSwap{}, \opInsert{}, and \opDelete{}, and verify the resulting samples for visual, acoustic, and synchronization fidelity. In Phase 3, we construct cross-pair appearance and voice references through identity retrieval and augmentation, ensuring that the references preserve the desired identity and timbre while differing in pose, scene, and linguistic content. Additionally, structured editing instructions are annotated using Qwen3.5~\citep{QwenTeam2026Qwen35} (detailed in Appendix~\ref{app:instruction-gen}) to guide cross-modal editing intents. To guarantee dataset fidelity, a multi-stage verification suite (detailed in Appendix~\ref{app:quality-verification}) rigorously filters out visual artifacts, lip-speech misalignments, and identity-timbre mismatches throughout the pipeline. The resulting \datasetname{} contains 100K curated audio-visual editing pairs, including 40K \opSwap{} pairs, over 35K \opInsert{} pairs, and over 20K \opDelete{} pairs.

% \subsubsection{Source Corpus and Filtering}
% We collect videos featuring human subjects from films, television programs, and short-form video platforms. We filter the collection along multiple dimensions, including file specifications (duration, resolution, frame rate, and audio sampling rate), visual quality, temporal continuity and motion, visual contamination, suitability of the person and speaking scene, audio quality, audio-visual consistency, and content safety. We then stratify the retained clips and balance their distributions across key attributes. Further implementation details are provided in Appendix~\ref{app:dataset-profile}.

\subsubsection{Edit Pair Synthesis}
We construct high-quality editing pairs with aligned audio and video through a multi-stage synthesis and filtering pipeline (Fig.~\ref{fig:data}, Stage 2). The pipeline has two branches: \opSwap{}, and the complementary \opDelete{}/\opInsert{} operations. The resulting editing pairs support both distillation from synthetic targets and reconstruction of real targets.

% 8.5--这两个是放 Edit Pair Synthesis 下面

\paragraph{Swapping.}
Person swapping replaces the appearance and voice timbre of a target person in the source video with those of a reference individual. It should faithfully preserves the original motion and poses, speech timing, and background audio.
For video synthesis (Stage 2.A), we employ Wan 2.2 Animate~\citep{Cheng2025WanAnimate} with filtered source video clips serving as motion and pose drivers, combining them with reference facial images to synthesize videos with replaced appearances (\textit{Person-Swapped Video}).
For audio synthesis (Stage 2.B), we first utilize BS-RoFormer~\citep{Lu2024BSRoFormer} to precisely separate the source audio into clean speech and background noise/sound effect tracks. Subsequently, Seed-VC V2~\citep{Liu2024SeedVC} performs voice conversion using the isolated speech for content and prosody guidance with the reference timbre. The cloned speech is then remixed with background audio, yielding a target track synchronized with the edited video.

\paragraph{Removal \& Insertion.}
Removal and insertion are inverse audio-visual editing operations.
For video-level erasing, we track the target person in the source video using SAM 2~\citep{ravi2025sam} to generate frame-wise binary masks, which are then fed into MiniMax-Remover~\citep{zi2026minimax} for inpainting to yield a background video with the target person removed (\textit{Person-Removed Video}).
For audio-level separation, we similarly strip the target voice via BS-RoFormer, retaining only ambient background sounds and audio effects (or silence if no background audio exists).
From the resulting video and audio, we construct bidirectional editing pairs:
1) \textbf{Removal}: The original audio-visual clip serves as the source, while the person-removed video and background audio serve as the target;
2) \textbf{Insertion}: We reverse the pair and provide reference images and audio of the person to be inserted, using the original clip as the target. Thus, insertion is supervised by real, synchronized audio-visual targets.

\subsubsection{Cross-Pair Reference Construction}
%References cropped directly from the paired target may share its pose, scene, or linguistic content, allowing the model to exploit a copy-and-paste shortcut instead of learning identity and timbre. 
For reference-conditioned backward (using synthetic clips as inputs and real clips as targets) reconstruction, we need references for the original person. Simply cropping the real target would expose its pose, scene, or speech content, encouraging copy-and-paste shortcuts.
Therefore, we construct cross-pair visual and audio references that match the target's identity and clothing but differ in background, pose, expression.
For visual references, we sample and crop one face per second from each video. We then compute face embeddings and use them to index the crops in an identity retrieval database. We retrieve candidate references of the same identity and further verify them through speaker-timbre similarity. After suppressing background sounds and sound effects, the corresponding speech segments serve as candidate audio references. Qwen3.5 rejects neighboring excerpts from the same continuous scene and candidates with inconsistent clothing, accessories, or hairstyles, which commonly arise when the same actor appears in different productions. We complement retrieved references with generative augmentation via Qwen-Image-Edit~\citep{Wu2025QwenImage} and CosyVoice 3~\citep{Du2024CosyVoice}. For visual augmentation, an image editor swaps the subject's background with samples from a background library while introducing subtle variations in expression and pose. For audio augmentation, CosyVoice 3 synthesizes new utterances from a text corpus using clean speech segments as voice prompts. All synthesized references undergo a final quality check (detailed in Appendix~\ref{app:quality-verification}).

\subsection{Framework of \methodname{}}

With the paired data in place, we build our editor on a joint audio--video generation backbone, so that editing inherits a strong prior on how speech, lip motion, and ambient sound co-occur instead of having to relearn cross-modal consistency from the editing pairs alone. Our design relies on only three properties shared across this family of models~\citep{Low2025Ovi,HaCohen2026LTX2,SkyReelsTeam2026SkyReelsV4}: separate video and audio VAEs map the two streams into their own latent spaces; two Diffusion Transformer (DiT) branches are coupled by bidirectional cross-modal attention, so that each modality conditions on the other at every block; and a shared frozen text encoder supplies the semantic embedding of the editing instruction. Because we assume nothing beyond these properties, our method is not tied to a particular backbone. We instantiate it with Ovi~\citep{Low2025Ovi} and adapt the conditional DiT blocks to source-conditioned editing.
\label{sec:framework}
\begin{figure*}[t]
    \centering
    \includegraphics[width=0.90\linewidth]{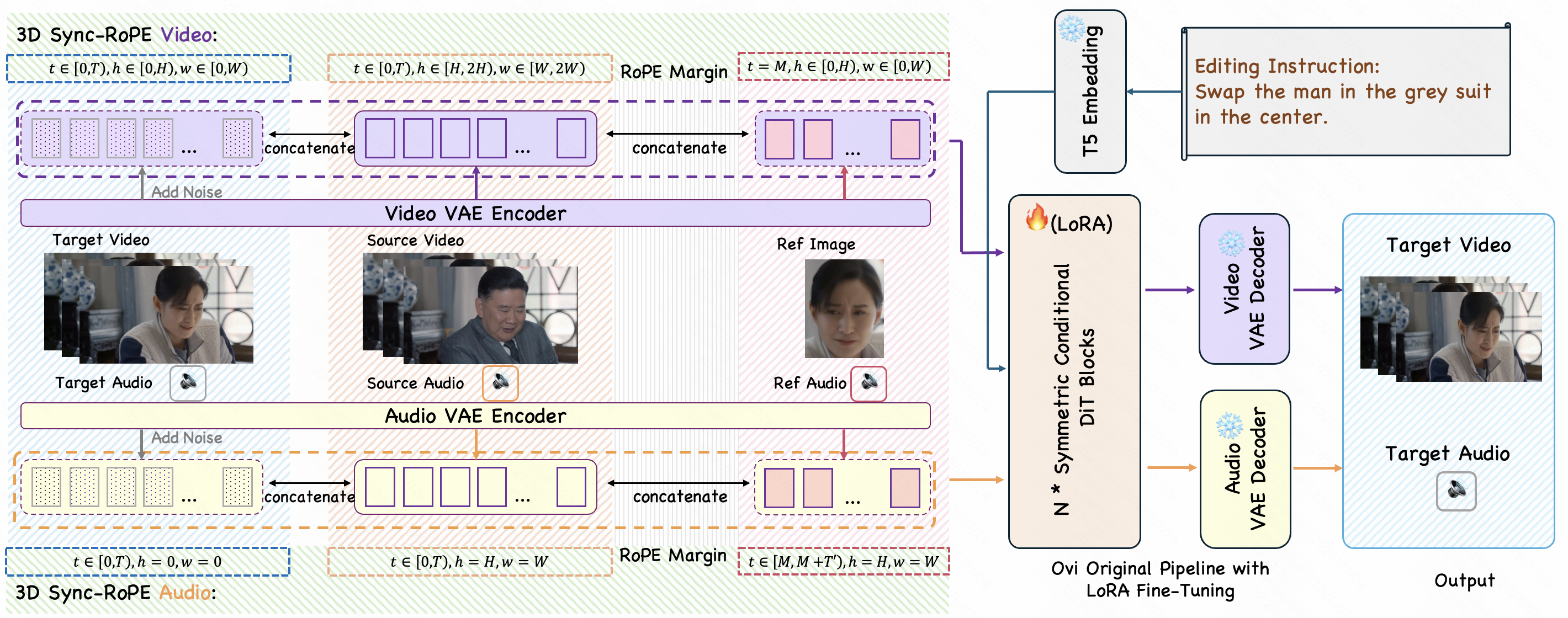}
    \caption{Overview of the \methodname{} framework.
	% Source context, reference inputs, and noisy target streams are concatenated for joint denoising.
	To prevent token ambiguity, our \syncrope{} introduces role-specific 3D coordinate offsets that explicitly distinguish source, target, and reference tokens while maintaining temporal correspondence along a shared timeline.
	%LoRA fine-tuning is applied exclusively to the conditional DiT blocks.
    }
    \label{fig:pipeline}
\end{figure*}

\subsubsection{Source- and Reference-Conditioned Editing}
% 图注  % 基于MMDIT架构
% 2.输入 怎么变输出（数据流走向的文字描述
% （都放在图里讲）
As illustrated in Fig.~\ref{fig:pipeline}, the source and reference inputs are encoded as clean context, whereas the target video and audio streams are initialized from noise. Within each modality, the target, source, and reference tokens are concatenated and jointly processed by the coupled DiT backbones. After joint denoising, the video and audio decoders produce the edited pair.

% \TODO{补 LoRA：图里画了 LoRA 微调，正文未提；说明哪些模块可训练。并明确"where to edit"由 instruction 加全局上下文决定，回收 intro 的 mask trilemma 论证。}

\subsubsection{\syncrope}
% 几句话说清楚如何区分source target。 尽管 Ovi/Ltx已经有了音视频同步Rope，但是在编辑的时候直接用会导致source target区分不清。
Inspired by DreamID-Omni~\citep{Guo2026DreamIDOmni}, we design a \syncrope{} layout that distinguishes the roles of target, source, and reference tokens without disrupting source--target temporal correspondence. Both streams share the same three coordinate axes, but each modality interprets them according to its own structure. Let $T$, $H$, and $W$ denote the temporal, height, and width dimensions of the video latent grid. We assign the target and source video tokens to
% \begin{equation}
% \begin{aligned}
% \Omega_{\mathrm{tar}}^v
%     &= [0,T)\times[0,H)\times[0,W),\\
% \Omega_{\mathrm{src}}^v
%     &= [0,T)\times[H,2H)\times[W,2W).
% \end{aligned}
% \label{eq:video-rope-layout}
% \end{equation}
\begin{equation}
\begin{aligned}
\Omega_{\mathrm{tar}}^v &= [0,T)\times[0,H)\times[0,W), \\
\Omega_{\mathrm{src}}^v &= [0,T)\times[H,2H)\times[W,2W).
\end{aligned}
\label{eq:video-rope-layout}
\end{equation}
Thus, corresponding source and target frames share the same temporal coordinates, while offsets along both spatial axes prevent their tokens from being interpreted as the same latent canvas. For an optional reference image, we use
\begin{equation}
    \Omega_{\mathrm{ref}}^v
    = \{M\}\times[0,H)\times[0,W),
\end{equation}
where the RoPE margin $M$ separates reference information from the source and target sequences.

% We apply the same principle to the audio stream. Let $T_a$ be the number of audio latent tokens. Since audio has no intrinsic spatial grid, we use the two auxiliary RoPE axes to encode token roles:
We apply the same principle to the audio stream. Let $T_a$ be the number of audio latent tokens; following Ovi~\citep{Low2025Ovi}, we scale their temporal RoPE frequencies by $\gamma=T/T_a$, such that $\gamma T_a=T$ and the audio and video streams span the same temporal RoPE range. Since audio has no intrinsic spatial grid, we use the two auxiliary RoPE axes to encode token roles:
% \begin{equation}
% \begin{aligned}
% \Omega_{\mathrm{tar}}^a
%     &= [0,T_a)\times\{0\}\times\{0\},\\
% \Omega_{\mathrm{src}}^a
%     &= [0,T_a)\times\{H\}\times\{W\},\\
% \Omega_{\mathrm{ref}}^a
%     &= [M,M+T_a)\times\{H\}\times\{W\}.
% \end{aligned}
% \label{eq:audio-rope-layout}
% \end{equation}
% \begin{equation}
% \Omega_{\mathrm{tar}}^a = [0,T_a)\times\{0\}\times\{0\}, \quad \Omega_{\mathrm{src}}^a = [0,T_a)\times\{H\}\times\{W\}, \quad \Omega_{\mathrm{ref}}^a = [M,M+T_a)\times\{H\}\times\{W\}.
% \label{eq:audio-rope-layout}
% \end{equation}
\begin{equation}
\begin{aligned}
\Omega_{\mathrm{tar}}^a &= [0,T_a)\!\times\!\{0\}\!\times\!\{0\}, \\
\Omega_{\mathrm{src}}^a &= [0,T_a)\!\times\!\{H\}\!\times\!\{W\}, \\
\Omega_{\mathrm{ref}}^a &= [M,M+T_a)\!\times\!\{0\}\!\times\!\{0\}
\end{aligned}
\label{eq:audio-rope-layout}
\end{equation}
The source and target audio tokens therefore remain temporally aligned, while the reference audio is isolated by the same margin. We retain Ovi's~\citep{Low2025Ovi} temporal RoPE scaling between the audio and video branches so that their different token rates correspond to the same physical timeline.
%有一点长，
% 正文说 source 与 target 共享时间坐标 [0,T)，但 pipeline.png 里 source 标的是 t∈[T,2T)，图文矛盾

\subsection{Progressive Training} % 渐进学习，不要一个training概括
\label{sec:training}
% 为了更高效利用我们的合成数据训练我们的管线JAVE，并进一步图片合成数据上限，我们采取了双向（、多阶段训练策略。
\subsubsection{Bidirectional Supervision}
% 换个说法，看看VSR他们怎么说的
Training only in the forward direction (from a real source to a synthetic target) would primarily distill the capabilities and artifacts of the data-synthesis pipeline. We therefore additionally reverse each editing-pair: the synthesized sample is used as the source and its real counterpart as the supervision target. This backward direction exposes the model to real target distributions and improves generation quality. We train primarily on backward pairs, supplemented by forward pairs. Forward supervision teaches the intended edits from real source videos without compromising the pretrained generation prior, whereas backward supervision improves output fidelity without encouraging the model to exploit residual synthesis artifacts as shortcuts. Our stringent filtering makes the synthetic targets used in forward training and the synthetic inputs used in backward training closely approximate real audio--video data.

% 如果我们直接将合成数据用语训练（Forward Training Paradigm），那么相当于只是蒸馏数据合成管线的能力。
% 相反地，我们采用在训练中加入反向训练，也就是以合成数据做输入。并在后续试验中平衡两者的比例，确保：
% 正向训练不会让模型生成质量变差，但是能让模型学会如何从原始视频编辑
% 反向训练提高模型生成质量，但不会迫使模型根据潜在的编辑痕迹hack
% 由于对构造视频质量的筛选和检测，我们确保正向训练的监督数据和反向训练的输入数据接近真实音-视频。

\subsubsection{Multi-Stage Training Curriculum}
\leavevmode\par
\myparagraph{Stage 0: Audio Warm-up.}
% We first isolate the audio branch and train it on audio-only synthetic data for TTS, voice conversion, and speaker-timbre disentanglement. This warm-up is critical for mitigating the severe audio--video token imbalance and substantially accelerates subsequent joint optimization. The warmed audio editor can be integrated into the joint model with only a short adaptation phase in Stage 1, whose compute is approximately 10\% of the subsequent training cost.
We first isolate the audio branch and train it on audio-only synthetic data for TTS, voice conversion, and speaker-timbre disentanglement. This warm-up is critical for mitigating the severe audio--video token imbalance and substantially accelerates subsequent joint optimization, allowing the audio editor to be efficiently integrated into the joint model.

\myparagraph{Stage 1: Bidirectional Joint Adaptation.}
% We train with both forward and backward pseudo-pairs. For the first 2K steps, we freeze the audio branch and optimize the video and cross-modal fusion modules on joint audio--video data. We then unfreeze the audio branch and jointly optimize all trainable modules for another 2K steps. At this point, the model acquires the three editing operations, although fine-grained defects remain, including deletion artifacts and temporal inconsistencies in inserted or swapped subjects.
We train with both forward and backward directions. We first freeze the audio branch and optimize the video and cross-modal fusion modules on joint audio-video data, and subsequently unfreeze the audio branch to jointly optimize all trainable modules. This enables the model to acquire all three editing operations within a unified framework.

\myparagraph{Stage 2: Quality Refinement.}
We select the highest-quality subset from each data category based on VBench~\citep{huang2024vbench}, SyncNet~\citep{Chung2017SyncNet}, and other quality scores, retaining only samples with retrieved cross-pair references for continued backward training. This quality-focused stage improves local details and reduces the residual artifacts from Stage 1.

% 待改：图注写的是 two-stage，正文是三个 stage 而实际分四段，图文需一致。
% 待改：数据图用 Stage 1/2.A/2.B/3，与本节 Stage 0/1/2 撞车，数据侧改成 Step。

% stage0将音频模型单独抽取出来，针对音频构造数据训练tts vc 音色分离能力。
% 我们发现这个训练是至关重要的————完美加速了模型训练效率，解决了音/视频模态token不均衡的问题。
% 并且少量的stage1预热就可以适配这个阶段训练好的纯音频编辑模型，需要的计算量大约是后续训练的10%。
% 
% stage1使用正向/反向训练
% 这个阶段首先将视频和fusion模块与音频做适配：冻结音频模块权重，在音-视频联合数据上训练2k steps，之后打开音频模块权重继续训练2k steps
% 在这个阶段模型基本上学会三种任务，但是部分细节较为粗糙，例如可能删除时产生伪影，或者插入/替换的人物时序一致性不是很好
% 
% stage2使用反向训练、高质量数据约束，我们对构造数据时每个分类仅使用stage1中根据vbench syncnet等打分后最好的1/2数据
% 这个阶段继续训练2k steps。解决stage1潜在的细节建模问题

% \TODO{全文目前没有任何 loss，需补 flow matching 训练目标；以及推理设置（CFG scale、采样步数、instruction 的具体形式）。}

\section{Experiments}
\subsection{Experimental Setup}
\myparagraph{Implementation Details.}
We fine-tune Ovi~\citep{Low2025Ovi} using LoRA~\citep{hu2021lora}; trainable modules are detailed in Appendix~\ref{app:training-details}. Training videos are standardized to a resolution of $480\times832$ at 24~fps with a duration of 5.04~seconds (121 frames), aligning with Ovi's temporal grid. Audio is sampled at 16~kHz. For \syncrope{}, we set the shared RoPE margin to $M=350$ for both the video and audio branches. Following the curriculum in Sec.~\ref{sec:training}:
In Stage 0, we train the audio-only LoRA parameters with a batch size of 512 for 2k steps. In Stage 1, we train on audio-video pairs with a batch size of 64, running 2k steps with frozen audio LoRA parameters followed by 2k steps of joint optimization across all LoRA parameters.
In Stage 2, we continue backward training on the top-50\% quality-refined subset with a batch size of 64 for 2k steps.

\begin{figure*}[t]
	\centering
	\includegraphics[width=0.88\textwidth]{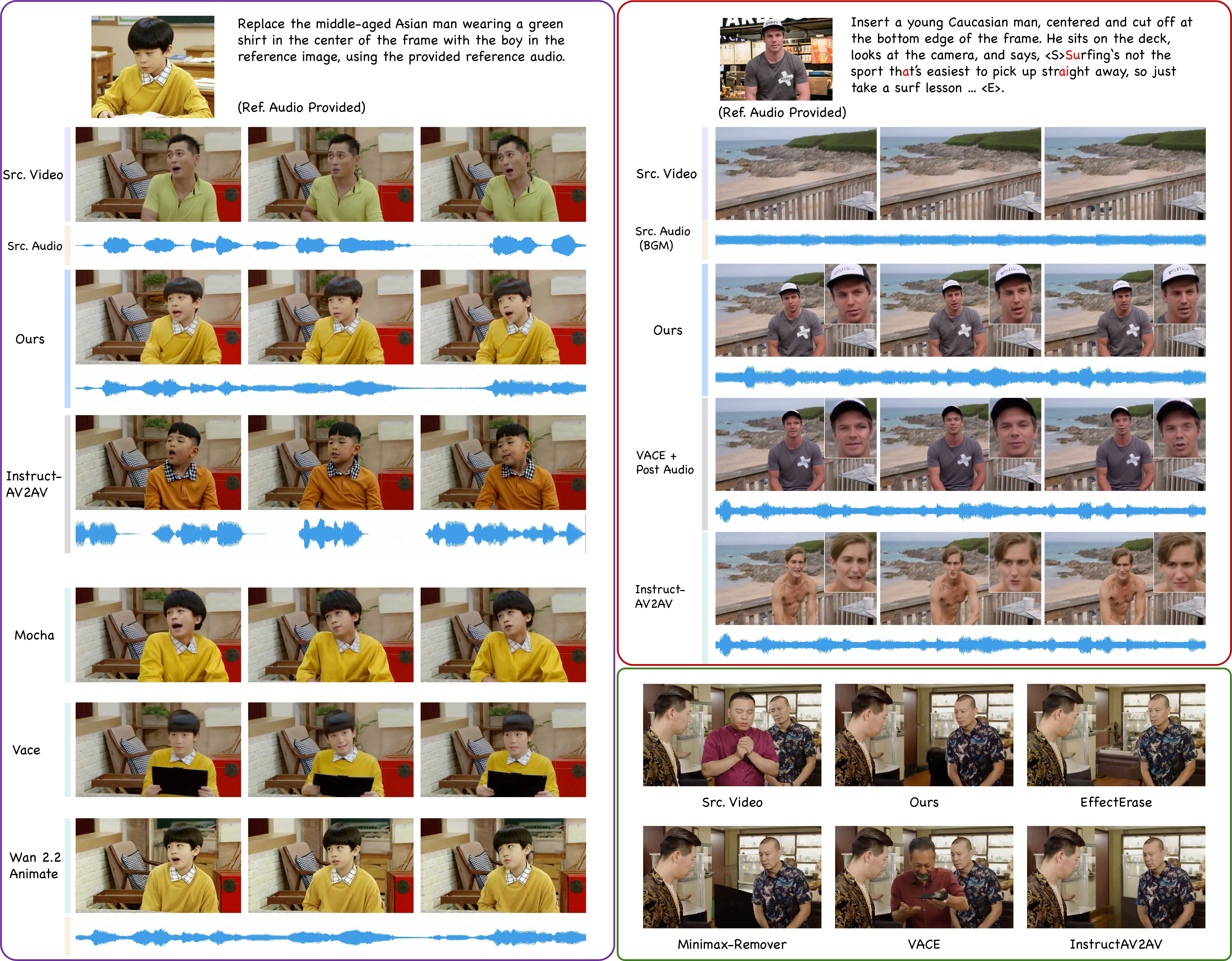}
	\caption{
    %Qualitative comparisons across person swapping(purple), insertion(red), and removal(green).
    Qualitative comparisons across person \textcolor[HTML]{68339A}{swapping}, \textcolor[HTML]{AF2519}{insertion}, and \textcolor[HTML]{4C7C31}{removal}. Please zoom in for more details.
    }
	\label{fig:main-results}
\end{figure*}

\myparagraph{Benchmark Evaluation.}
Publicly available benchmarks closest to our setting remain fragmented. InstructAV2AV and JAVEdit provide multi-operation evaluation sets based on text instructions, without external appearance or voice references~\citep{Zheng2026InstructAV2AV,Chen2026JAVEdit}. IDBench-Omni provides reference-conditioned evaluation, but its released source-video editing set comprises only 25 full-human-swap cases and 25 face-swap cases, with target dialogue specified by text prompts; person insertion and deletion into existing source videos are not evaluated~\citep{Guo2026DreamIDOmni}.

To address these gaps, we construct \datasetname-Bench, comprising held-out video clips strictly disjoint from our training set. Specifically, \datasetname-Bench contains 64 carefully curated test samples for each of the three editing tasks, yielding 192 test cases in total. Models receive real-world audio-visual clips as source inputs, paired with reference images and audio as conditioning signals.

\myparagraph{Baselines.}
We compare \methodname{} with two categories of baselines: separate audio-video editing pipelines and the text-driven joint audio-video editing model InstructAV2AV~\citep{Zheng2026InstructAV2AV}. 
For separate audio--video editing, each visual editor is paired with the corresponding task-specific audio editing pipeline. Specifically, we use VACE~\citep{Jiang2025VACE} as the general-purpose visual baseline across all three tasks: \opSwap{}, \opDelete{}, and \opInsert{}. In addition, for \opSwap{}, we evaluate Wan-Animate~\citep{Cheng2025WanAnimate} and MoCha~\citep{Xu2026MoCha}; for \opDelete{}, we evaluate MiniMax-Remover~\citep{zi2026minimax} and EffectErase~\citep{fu2026effecterase}. To benchmark against end-to-end joint editing, we further compare with InstructAV2AV. Since InstructAV2AV is conditioned solely on text instructions and does not natively accept external visual or vocal references, we convert the reference subject's salient appearance and vocal attributes into detailed text descriptions and append them to the instruction prompt.

\myparagraph{Evaluation Metrics.}
We evaluate the edited results across five dimensions:
(1) Visual quality: five VBench dimensions~\citep{huang2024vbench}, including subject consistency, background consistency, temporal flickering, motion smoothness, and imaging quality;
(2) Acoustic quality: AudioBox PQ~\citep{tjandra2025meta} ($\uparrow$) measuring the acoustic quality of generated speech;
(3) Audio-visual synchronization: Sync-C ($\uparrow$, confidence) and Sync-D ($\downarrow$, feature distance) evaluated via SyncNet~\citep{Chung2017SyncNet};
(4) Reference fidelity: ArcFace~\citep{deng2019arcface} ($\uparrow$) for facial identity similarity, CAM++ ~\citep{wang2023cam++} ($\uparrow$) for speaker timbre similarity, and CLIP-I~\citep{radford2021learning} ($\uparrow$) for image-level subject similarity;
(5) Background fidelity: PSNR ($\uparrow$) and LPIPS~\citep{zhang2018unreasonable} ($\downarrow$) computed over the background region. For \opSwap{} and \opDelete{}, the background is obtained by segmenting the edited subject via SAM~2~\citep{ravi2025sam} followed by a moderate mask dilation. For \opInsert{}, the inserted subject is segmented from the edited output, and the unmasked region is taken as the background.

\begin{table*}[t]
\centering
\caption{Quantitative comparison on \datasetname-Bench across three editing tasks. Reconstruction metrics are computed over the background region. Best results are shown in bold. ``+PA'' denotes cascaded baselines paired with our audio post-processing pipeline.}
\label{tab:jave-other-metrics}
	\begingroup
	\scriptsize
		\setlength{\tabcolsep}{1.5pt}
	\renewcommand{\arraystretch}{0.95}
		\begin{tabular}{@{\hskip 0pt}l*{13}{c}@{\hskip 0pt}}
		\toprule
		& \multicolumn{5}{c}{\textbf{VBench (\%)}}
		& \multicolumn{3}{c}{\textbf{Audio Quality \& AV Sync}}
		& \multicolumn{3}{c}{\textbf{Ref Similarity}}
		& \multicolumn{2}{c}{\textbf{BG Fidelity}} \\
		\cmidrule(lr){2-6}\cmidrule(lr){7-9}\cmidrule(lr){10-12}\cmidrule(l){13-14}
		\textbf{Task / Method}
		& \shortstack{Subj.\\Cons.} $\uparrow$
		& \shortstack{BG\\Cons.} $\uparrow$
		& \shortstack{Temp.\\Flick.} $\uparrow$
		& \shortstack{Motion\\Sm.} $\uparrow$
		& \shortstack{Imaging\\Q.} $\uparrow$
		& PQ $\uparrow$ & Sync-C $\uparrow$ & Sync-D $\downarrow$
		& CAM++ $\uparrow$ & Arc $\uparrow$ & CLIP $\uparrow$
		& PSNR $\uparrow$ & LPIPS $\downarrow$ \\
		\midrule

% Swap
\multicolumn{14}{@{}l}{\textbf{Swap}} \\
VACE+PA & 95.26 & 94.30 & 96.48 & 98.11 & 68.13 & 6.586 & 4.86 & 9.29 & 0.665 & 0.614 & 0.705 & \best{35.37} & \best{0.0481} \\
Wan2.2+PA & 94.09 & \best{96.12} & 97.76 & 97.71 & 66.41 & 6.586 & 5.02 & 8.76 & 0.665 & 0.634 & 0.710 & 33.27 & 0.0505 \\
MoCha+PA & 96.52 & 93.16 & \best{97.97} & 97.30 & 65.11 & 6.586 & 5.05 & 8.72 & 0.665 & 0.639 & \best{0.712} & 32.42 & 0.0584 \\
InstructAV2AV & 95.86 & 93.88 & 95.56 & 98.10 & 66.60 & 6.255 & 3.87 & 9.65 & -- & -- & -- & 31.84 & 0.0640 \\
\rowcolor{gray!15}
\textbf{Ours} & \best{97.35} & 95.69 & 97.67 & \best{98.54} & \best{69.74} & \best{6.709} & \best{5.39} & \best{8.53} & \best{0.682} & \best{0.647} & \best{0.712} & 34.62 & 0.0549 \\
\midrule

% Delete
\multicolumn{14}{@{}l}{\textbf{Delete}} \\
VACE+PA & 93.06 & 92.72 & 95.61 & 97.59 & 65.64 & \best{6.813} & -- & -- & -- & -- & -- & 29.94 & 0.1129 \\
MiniMax-Remover+PA & 97.39 & 95.90 & 97.97 & 97.98 & 59.07 & \best{6.813} & -- & -- & -- & -- & -- & 27.99 & 0.1317 \\
Effect Erase+PA & 97.69 & 96.24 & 98.04 & 96.20 & 62.65 & \best{6.813} & -- & -- & -- & -- & -- & 31.27 & 0.1212 \\
InstructAV2AV & 96.53 & 93.82 & 96.32 & 97.56 & 60.63 & 6.573 & -- & -- & -- & -- & -- & 29.74 & 0.0991 \\
\rowcolor{gray!15}
\textbf{Ours} & \best{98.49} & \best{96.64} & \best{98.12} & \best{98.43} & \best{67.15} & 6.810 & -- & -- & -- & -- & -- & \best{31.74} & \best{0.0926} \\
\midrule

% Insert
\multicolumn{14}{@{}l}{\textbf{Insert}} \\
VACE+PA & 96.04 & 94.71 & 97.61 & 97.70 & 68.20 & 6.384 & 3.93 & 9.73 & 0.638 & 0.614 & \best{0.705} & 28.66 & 0.1271 \\
InstructAV2AV & 95.71 & 93.52 & \best{98.10} & \best{98.63} & 63.72 & 6.090 & 3.84 & 10.60 & -- & -- & -- & \best{29.38} & \best{0.0950} \\
\rowcolor{gray!15}
\textbf{Ours} & \best{98.36} & \best{96.87} & 98.05 & 98.44 & \best{72.82} & \best{6.491} & \best{4.43} & \best{8.83} & \best{0.651} & \best{0.630} & 0.698 & 27.70 & 0.1023 \\
		\bottomrule
	\end{tabular}
	\endgroup
\end{table*}

\begin{table*}[t]
\centering
\caption{Ablation study on \datasetname-Bench across tasks. Best results are shown in bold.}
\label{tab:ablation-jave-bench}
	\begingroup
	\scriptsize
% 	\setlength{\tabcolsep}{1.2pt}
% \renewcommand{\arraystretch}{0.95}
% 	\begin{tabular}{@{}l*{8}{c}@{}}
	\setlength{\tabcolsep}{5pt}
\renewcommand{\arraystretch}{0.95}
	\begin{tabular}{@{}l@{\hspace{3pt}}*{5}{c}@{\hspace{3pt}}*{3}{c}@{}}
		\toprule
		& \multicolumn{5}{c}{\textbf{VBench (\%)}}
		& \multicolumn{3}{c}{\textbf{Audio Quality \& AV Sync}} \\
		\cmidrule(lr){2-6}\cmidrule(l){7-9}
		\textbf{Task / Variant}
		& \shortstack{Subj.\\Cons.} $\uparrow$
		& \shortstack{BG\\Cons.} $\uparrow$
		& \shortstack{Temp.\\Flick.} $\uparrow$
		& \shortstack{Motion\\Sm.} $\uparrow$
		& \shortstack{Imaging\\Q.} $\uparrow$
		& PQ $\uparrow$ & Sync-C $\uparrow$ & Sync-D $\downarrow$ \\
		\midrule

% Swap
\multicolumn{9}{@{}l}{\textbf{Swap}} \\
w/o Audio Pretrain & 96.60 & 95.34 & \best{97.94} & 97.95 & 68.58 & 6.107 & 4.71 & 9.25 \\
w/o Multistage & 95.90 & 94.57 & 97.28 & 98.21 & 67.88 & 6.580 & 5.28 & 8.66 \\
\rowcolor{gray!15}
\textbf{Ours} & \best{97.35} & \best{95.69} & 97.67 & \best{98.54} & \best{69.74} & \best{6.709} & \best{5.39} & \best{8.53} \\
\midrule

% Delete
\multicolumn{9}{@{}l}{\textbf{Delete}} \\
w/o Audio Pretrain & 97.87 & 96.20 & 97.93 & \best{98.44} & 65.57 & 6.304 & -- & -- \\
w/o Multistage & 96.57 & 95.08 & 97.61 & 98.13 & 63.27 & 6.704 & -- & -- \\
\rowcolor{gray!15}
\textbf{Ours} & \best{98.49} & \best{96.64} & \best{98.12} & 98.43 & \best{67.15} & \best{6.810} & -- & -- \\
\midrule

% Insert
\multicolumn{9}{@{}l}{\textbf{Insert}} \\
w/o Audio Pretrain & 96.75 & 94.95 & 98.37 & 97.97 & 68.21 & 5.980 & 3.54 & 9.75 \\
w/o Multistage & 96.69 & 95.21 & \best{98.47} & 97.60 & 68.22 & 6.320 & 4.15 & 9.18 \\
\rowcolor{gray!15}
\textbf{Ours} & \best{98.36} & \best{96.87} & 98.05 & \best{98.44} & \best{72.82} & \best{6.491} & \best{4.43} & \best{8.83} \\
		\bottomrule
	\end{tabular}
	\endgroup
\end{table*}

\subsection{Main Results}
Table~\ref{tab:jave-other-metrics} summarizes the quantitative comparisons across the three editing tasks, and Figure~\ref{fig:main-results} shows representative qualitative examples. Overall, \methodname{} delivers competitive visual and acoustic quality, strong audio--visual synchronization, and high reference fidelity.

\myparagraph{Swapping.}
In Figure~\ref{fig:main-results}, \methodname{} seamlessly replaces the target subject with the reference child. In contrast, VACE exhibits lip-speech misalignment, Wan-Animate alters the background layout, and MoCha degrades facial fidelity. InstructAV2AV modifies appearance via text prompts, but cannot recover the specific reference identity. Quantitatively, \methodname{} achieves the highest identity fidelity (ArcFace $0.682$, CAM++ $0.647$) and best audio--visual synchronization (Sync-C $5.39$, Sync-D $8.53$).

\myparagraph{Insertion.}
For person insertion, \methodname{} generates a new speaker with precise lip synchronization on key syllables and clear facial details. VACE suffers from blurred eyes and poor lip sync, while InstructAV2AV produces an unnatural pose and disobey the prompt. In quantitative evaluation, \methodname{} substantially improves audio--visual alignment (Sync-C $4.43$, Sync-D $8.83$ versus $3.93$, $9.73$ for VACE+PA) and achieves a higher imaging quality score ($72.82$).

\myparagraph{Removal.}
For subject removal, \methodname{} naturally infills the vacated region without explicit masks. EffectErase generates discordant background textures, MiniMax-Remover leaves black artifacts, VACE fails to delete the target, and InstructAV2AV distorts the adjacent person while blurring the background. \methodname{} achieves both the best background preservation (PSNR $31.74$, LPIPS $0.0926$) and the highest imaging quality ($67.15$).

\subsection{Ablation Study and Analysis}
We conduct ablation studies on \datasetname-Bench to validate our core training strategies, additional ablation of \syncrope{} is provided in Appendix~\ref{app:additional-ablations}.

\myparagraph{Effect of Audio Warm-up (Stage 0).}
To verify the necessity of isolated audio warm-up, we compare our full model against a variant that skips Stage 0 and directly initiates joint training from Stage 1 through Stage 2 (with extra training in Stage 1 to ensure equal total training iterations). As demonstrated in our experiments, omitting Stage 0 causes severe cross-modal token imbalance and leads to degraded speech quality (AudioBox PQ) and poorer AV synchronization (Sync-C), confirming that preliminary audio warm-up provides crucial acoustic priors for joint optimization.

\myparagraph{Effect of Multi-Stage Refinement (Stage 2).}
We evaluate the impact of Stage 2 quality refinement by comparing against a variant that continues training exclusively on Stage 1 data for an equal number of iterations without quality filtering. As shown in Table~\ref{tab:ablation-jave-bench}, omitting Stage 2 leads to a performance degradation, particularly in visual quality and consistency metrics evaluated by VBench.

\section{Conclusion}
We present \methodname{}, an end-to-end framework for mask-free human-centric audio-visual editing that unifies person \opSwap{}, \opInsert{}, and \opDelete{} within a single generative process while preserving external appearance and voice reference control. To overcome data scarcity, we construct \datasetname{}, a 100K-pair audio-visual editing dataset with multimodal references, and train the model primarily with backward reconstruction to leverage authentic targets as supervision. Experimental results across all three tasks demonstrate that \methodname{} achieves stronger editing fidelity, temporal stability, and lip-speech synchronization than cascaded editing baselines.
%In this paper, we present \methodname{}, a unified end-to-end framework for mask-free human-centric audio-visual editing. \methodname{} jointly performs \opSwap{}, \opInsert{}, and \opDelete{} within a single generative process, preserving operation-irrelevant visual and acoustic context while maintaining lip--speech synchronization. To address the scarcity of naturally paired supervision, we construct a scalable pipeline that uses specialized editors as offline data factories to synthesize and verify complex editing pairs, together with cross-pair references that prevent content-copying shortcuts. Our source and reference conditioned joint architecture couples audio and video during denoising, while \syncrope{} distinguishes their target, source, and reference tokens without sacrificing source--target temporal correspondence. Finally, the progressive curriculum combines audio warm-up, bidirectional supervision, and quality refinement to reduce synthetic artifacts and improve joint editing fidelity. Experiments across all three operations demonstrate that \methodname{} yields more realistic, temporally stable, and synchronized results than cascaded editing baselines.

\subsection*{Reproducibility Statement}
To ensure the reproducibility of our results, all key architectural designs, mathematical formulations, and coordinate assignments of \syncrope{} are presented in Section~\ref{sec:framework} and Section~\ref{sec:training}. Detailed training hyperparameters, stage schedules, and optimization configurations are provided in Appendix~\ref{app:training-details}. The data curation workflow, filtering criteria, and benchmark protocols are described in Section~\ref{sec:data} and Appendix~\ref{app:dataset-profile}. The source code, evaluation benchmark (\datasetname-Bench), and model weights will be made publicly available for research purposes upon publication.

\subsection*{Ethics Statement}
This work explores human-centric audio-visual editing for creative content generation and multimedia post-production. We recognize the potential risks of identity manipulation, deepfake generation, and unauthorized voice cloning. To mitigate these risks, all video and audio datasets used in this research were gathered exclusively from publicly available sources for academic research purposes. We strictly oppose malicious uses of identity manipulation and advocate the integration of synthetic media detection frameworks and digital watermarking alongside generative models.

\subsection*{AI Use Statement}
In this work, generative foundation models and specialized tools were used as algorithmic components of our data curation and training pipeline: Qwen3.5 for instruction annotation and visual filtering, Wan-Animate and MiniMax-Remover for video synthesis, and BS-RoFormer, Seed-VC, and CosyVoice~3 for audio separation, voice conversion, and speech augmentation. Generative LLMs were also used solely for text polishing and proofreading of the manuscript. All synthesized datasets and final manuscript texts were carefully verified and audited by the authors. We take full responsibility for the content and artifacts presented in this paper.

\clearpage
\subsection*{Acknowledgments}
We thank Xiangyu Du and Ning Lei for data support, Da Xie and Tianpeng Lan for data and resource support, and Kling AI for its support.

{
\small
\bibliographystyle{ieeenat_fullname}
\bibliography{iclr2027_conference}
}

\iffalse
\appendix
\section{Appendix}
You may include other additional sections here.
\fi

\clearpage
\appendix
\section{Dataset Profile and Curation Details}
\label{app:dataset-profile}

\subsection{Source Corpus and Pre-Filtering}
\label{app:source-prefilter}
We collect videos featuring human subjects from films, television programs, and short-form video platforms. We filter the collection along multiple dimensions, including file specifications (duration, resolution, frame rate, and audio sampling rate), visual quality, temporal continuity and motion, visual contamination, suitability of the person and speaking scene, audio quality, audio-visual consistency, and content safety. We then stratify the retained clips and balance their distributions across key attributes.

\subsection{Editing Instruction Generation}
\label{app:instruction-gen}
To enable precise model comprehension of cross-modal editing intents, we leverage Qwen3.5~\citep{QwenTeam2026Qwen35} to jointly analyze source video frames, bounding boxes of edited/erased regions, and reference images, automatically generating structured instructions:
1) \textbf{Unique Referring Description}: Details the target person’s absolute and relative spatial position along with unique appearance traits, grounding the referring expression to prevent referent ambiguity during inference;
2) \textbf{Action Prompts}: concise editing directives (e.g., ``\textit{Replace the man in suit with the reference person}'' or ``\textit{Remove the speaker}'').
These two components are combined into a complete instruction via templates and polished using Qwen3.5 to ensure strong generalization across diverse user inputs in real-world inference scenarios.

For \opInsert{}, since the reference image and audio specify only the target person's identity (appearance and timbre) but not their action or speech content in the video, we additionally annotate this information within the instruction: actions are annotated by Qwen3.5, while speech content is primarily derived from subtitle OCR and pre-obtained scripts, with ASR serving as an auxiliary check. These annotations preserve the temporal synchronization between action and speech, alternately describing the action and the corresponding utterance at each moment rather than listing all speech content at once. All instructions strictly omit explicit identity or timbre descriptions present in the reference inputs (e.g., specific facial features or vocal traits), preventing the model from exploiting textual shortcuts to bypass reference encoders.

\subsection{Multi-Stage Quality Verification Suite}
\label{app:quality-verification}
To ensure pristine quality and fidelity across the synthesized dataset, we incorporate a rigorous quality verification suite throughout the entire synthesis workflow:
1) \textbf{Raw Data Pre-Filtering}: VBench~\citep{huang2024vbench} evaluates video aesthetics and temporal smoothness, while SyncNet~\citep{Chung2017SyncNet} filters out audio-visual synchronization offsets;
2) \textbf{Synthesis Quality Control}: Qwen3.5 checks the precision of SAM 2 masks and prompt bounding boxes, while ClearVoice~\citep{zhao2025clearervoice} performs denoising and enhancement on cloned speech;
3) \textbf{Cross-Pair Audit}: ArcFace verifies facial cosine similarity, CAM++ audits speaker identity consistency, and ASR combined with subtitle OCR cross-verifies spoken dialogue accuracy, ultimately yielding high-fidelity audio-visual editing pairs.

For generatively augmented cross-pair references, we apply additional modality-specific checks. Qwen3.5 verifies visual fidelity and the absence of visible artifacts, while ASR-based checks assess the linguistic correctness of synthesized speech. We then compute CLIP-based visual similarity~\citep{radford2021learning} and CAM++-based timbre similarity~\citep{wang2023cam++} between each synthesized reference and the original identity, retaining only samples with a CLIP score above 0.75 and a CAM++ score above 0.70.

\section{Training Details}
\label{app:training-details}

Table~\ref{tab:training-config} summarizes the data mixtures, optimization settings, and trainable modules for the curriculum in Sec.~\ref{sec:training}. All stages optimize rank-256 LoRA adapters; the pre-trained backbone weights, VAEs, and T5 text encoder remain frozen.

The filtered, camera-captured VOR subset~\citep{fu2026effecterase} provides real removal supervision in Stages 1 and 2. Quality refinement selects approximately the top 50\% of our constructed data pool as a whole while retaining this external subset.

\begin{table*}[t]
	\centering
	\caption{Stage-wise training configuration. Dataset sizes and task mixtures are approximate. The VOR subset is included in, not added to, the reported \opDelete{} counts.}
	\label{tab:training-config}
	\begingroup
	\small
	\setlength{\tabcolsep}{4pt}
	\renewcommand{\arraystretch}{1.18}
	\begin{tabular}{@{}>{\raggedright\arraybackslash}p{0.20\linewidth}>{\raggedright\arraybackslash}p{0.23\linewidth}>{\raggedright\arraybackslash}p{0.26\linewidth}>{\raggedright\arraybackslash}p{0.24\linewidth}@{}}
		\toprule
		\textbf{Configuration} & \textbf{Stage 0}\newline Audio Warm-up & \textbf{Stage 1}\newline Bidirectional Joint Adaptation & \textbf{Stage 2}\newline Quality Refinement \\
		\midrule
		Training data & 200K audio examples & 120K editing pairs & 70K editing pairs \\
		Task mixture & 30\% voice conversion;\newline 40\% text-to-speech;\newline 30\% vocal separation & 40K \opSwap{};\newline 37K \opInsert{};\newline 41K \opDelete{} & 25K \opSwap{};\newline 15K \opInsert{};\newline 30K \opDelete{} \\
        Reference source & -- & Retrieved + generated & Retrieved only \\
		VOR real-captured subset & -- & 20K \opDelete{} & 20K \opDelete{} \\
		\midrule
		Optimizer & AdamW & AdamW & AdamW \\
		Learning rate & $5\times10^{-5}$ & $1\times10^{-5}$ & $1\times10^{-5}$ \\
		LoRA rank & 256 & 256 & 256 \\
		Batch size & 512 & 64 & 64 \\
		Training steps & 2K & 2K + 2K & 2K \\
		Trainable LoRA modules & A only & First 2K: V + F;\newline next 2K: A + V + F & A + V + F \\
		\midrule
		Distributed training & DeepSpeed\newline ZeRO Stage 1 & DeepSpeed\newline ZeRO Stage 2 & DeepSpeed\newline ZeRO Stage 2 \\
		\bottomrule
	\end{tabular}
	\par\smallskip
	\begin{minipage}{\linewidth}
		\footnotesize
		A and V denote LoRA adapters in the audio and video DiT blocks, respectively; F denotes adapters in the bidirectional audio-video cross-modal fusion attention.
	\end{minipage}
	\endgroup
\end{table*}

\begin{table*}[t]
\centering
\caption{Ablation study of \syncrope{} on \datasetname-Bench across tasks. Best results are shown in bold.}
\label{tab:ablation-sync-rope}
	\begingroup
	\scriptsize
	\setlength{\tabcolsep}{5pt}
\renewcommand{\arraystretch}{0.95}
	\begin{tabular}{@{}l@{\hspace{3pt}}*{5}{c}@{\hspace{3pt}}*{3}{c}@{}}
		\toprule
		& \multicolumn{5}{c}{\textbf{VBench (\%)}}
		& \multicolumn{3}{c}{\textbf{Audio Quality \& AV Sync}} \\
		\cmidrule(lr){2-6}\cmidrule(l){7-9}
		\textbf{Task / Variant}
		& \shortstack{Subj.\\Cons.} $\uparrow$
		& \shortstack{BG\\Cons.} $\uparrow$
		& \shortstack{Temp.\\Flick.} $\uparrow$
		& \shortstack{Motion\\Sm.} $\uparrow$
		& \shortstack{Imaging\\Q.} $\uparrow$
		& PQ $\uparrow$ & Sync-C $\uparrow$ & Sync-D $\downarrow$ \\
		\midrule

\multicolumn{9}{@{}l}{\textbf{Swap}} \\
w/o Modality-Aware Sync-RoPE & 96.46 & \best{96.01} & 95.38 & 98.43 & 60.96 & 6.050 & 3.14 & 9.84 \\
\rowcolor{gray!15}
\textbf{Ours} & \best{97.35} & 95.69 & \best{97.67} & \best{98.54} & \best{69.74} & \best{6.709} & \best{5.39} & \best{8.53} \\
\midrule

\multicolumn{9}{@{}l}{\textbf{Delete}} \\
w/o Modality-Aware Sync-RoPE & 94.24 & 93.73 & 96.09 & 97.80 & 62.47 & 6.235 & -- & -- \\
\rowcolor{gray!15}
\textbf{Ours} & \best{98.49} & \best{96.64} & \best{98.12} & 98.43 & \best{67.15} & \best{6.810} & -- & -- \\
\midrule

\multicolumn{9}{@{}l}{\textbf{Insert}} \\
w/o Modality-Aware Sync-RoPE & 95.87 & 93.96 & 97.44 & \best{98.63} & 69.55 & 5.539 & 2.99 & 10.01 \\
\rowcolor{gray!15}
\textbf{Ours} & \best{98.36} & \best{96.87} & \best{98.05} & 98.44 & \best{72.82} & \best{6.491} & \best{4.43} & \best{8.83} \\
		\bottomrule
	\end{tabular}
	\endgroup
\end{table*}

\section{Additional Ablation Studies}
\label{app:additional-ablations}

\subsection{Effect of \syncrope{}}
We compare our full model with a conventional sequential RoPE variant that distinguishes token roles only by their positions along the first RoPE axis. For video, the target and source tokens occupy $[0,T)$ and $[T,2T)$ on this axis, respectively, and the reference image is placed at $2T$; all three use the same spatial coordinates $[0,H)\times[0,W)$. For audio, the target, source, and reference tokens occupy $[0,T_a)$, $[T_a,2T_a)$, and $[2T_a,3T_a)$, respectively, while sharing auxiliary coordinates $\{0\}\times\{0\}$. Thus, this variant removes both the role-specific offsets along the $H$ and $W$ axes and the fixed reference margin $M$ used by \syncrope{}. The token concatenation order, network architecture, training data, optimization schedule, and all other settings remain unchanged.

As shown in Table~\ref{tab:ablation-sync-rope}, replacing \syncrope{} with sequential RoPE consistently degrades imaging quality and AudioBox PQ across all three tasks. The full design improves imaging quality by 8.78, 4.68, and 3.27 points and AudioBox PQ by 0.659, 0.575, and 0.952 for \opSwap{}, \opDelete{}, and \opInsert{}, respectively. It also substantially improves audio-visual alignment for \opSwap{} and \opInsert{}, increasing Sync-C by 2.25 and 1.44 while reducing Sync-D by 1.31 and 1.18. Visual consistency generally improves as well, despite isolated exceptions in background consistency for \opSwap{} and motion smoothness for \opInsert{}. These results demonstrate that jointly preserving source--target temporal correspondence, separating token roles on the auxiliary axes, and isolating references with a fixed margin yields more reliable multimodal editing than sequential role encoding alone.

\end{document}